\documentclass[10pt,twocolumn,letterpaper]{article}

\usepackage{cvpr}              

\usepackage{graphicx}
\usepackage{amsmath}
\usepackage{amssymb}
\usepackage{booktabs}
\usepackage{xcolor}
\usepackage{multirow}
\usepackage{mathtools}
\usepackage{bm}

\usepackage[pagebackref,breaklinks,colorlinks]{hyperref}

\usepackage[capitalize]{cleveref}
\crefname{section}{Sec.}{Secs.}
\Crefname{section}{Section}{Sections}
\Crefname{table}{Table}{Tables}
\crefname{table}{Tab.}{Tabs.}

\newcommand{\methodname}{VQA-GNN\xspace}

\def\cvprPaperID{10971} 
\def\confName{ICCV}
\def\confYear{2023}

\begin{document}

\title{VQA-GNN: Reasoning with Multimodal Knowledge via Graph Neural Networks for Visual Question Answering}


\author{Yanan Wang${^1}$\thanks{~~~Work done while at Stanford University.}\quad Michihiro Yasunaga${^2}$\quad Hongyu Ren${^2}$\quad Shinya Wada${^1}$\quad Jure Leskovec${^2}$ \vspace{0.3em} \\
{${^1}$KDDI Research \quad ${^2}$Stanford University} \\
{{\tt \small${^1}$\{wa-yanan,sh-wada\}@kddi.com} \quad
{\tt \small ${^2}$\{myasu,hyren,jure\}@cs.stanford.edu}
}
}
\maketitle

\begin{abstract}
Visual question answering (VQA) requires systems to perform concept-level reasoning by unifying unstructured (\eg, the context in question and answer; ``QA context'') and structured (\eg, knowledge graph for the QA context and scene; ``concept graph") multimodal knowledge.  
Existing works typically combine a scene graph and a concept graph of the scene by connecting corresponding visual nodes and concept nodes, then incorporate the QA context representation to perform question answering. However, these methods only perform a unidirectional fusion from unstructured knowledge to structured knowledge, limiting their potential to capture joint reasoning over the heterogeneous modalities of knowledge. To perform more expressive reasoning, we propose VQA-GNN, a new VQA model that performs \textbf{bidirectional} fusion between unstructured and structured multimodal knowledge to obtain unified knowledge representations. Specifically, we inter-connect the scene graph and the concept graph through a super node that represents the QA context, and introduce a new multimodal GNN technique to perform inter-modal message passing for reasoning that mitigates representational gaps between modalities.
On two challenging VQA tasks (VCR and GQA), our method outperforms strong baseline VQA methods by \textbf{3.2\%} on VCR (Q-AR) and \textbf{4.6\%} on GQA, suggesting its strength in performing concept-level reasoning. Ablation studies further demonstrate the efficacy of the bidirectional fusion and multimodal GNN method in unifying unstructured and structured multimodal knowledge.
\end{abstract}
\section{Introduction}
\label{sec:intro}

\begin{figure}[ht]
\centering
\includegraphics[width=\columnwidth]{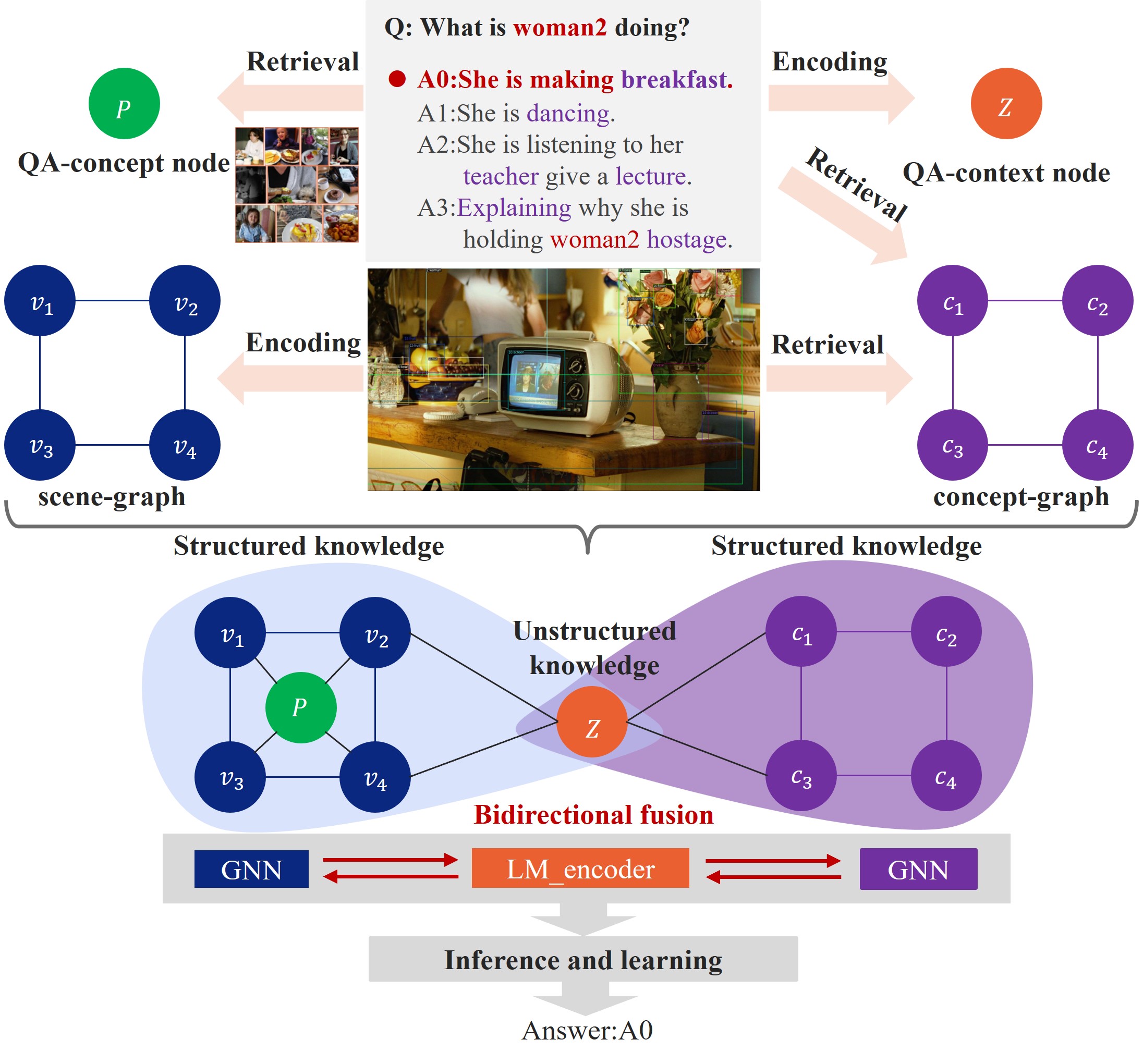}
\caption{Overview of \textit{\methodname}. Given an image and QA sentence, we obtain unstructured knowledge (\eg, QA-concept node p and QA-context node z) and structured knowledge (\eg, scene-graph and concept-graph), and then unify them to perform bidirectional fusion for visual question answering.} 
\label{fig:vqa_gnn_concept}
\end{figure}

The visual question answering (VQA) task aims to provide answers to questions about a visual scene. It is crucial in many real-world tasks including scene understanding, autonomous vehicles, search engines, and recommendation systems \cite{VQA,fukui2016multimodal,ben2017mutan,kim2018bilinear}. To solve VQA, systems need to perform concept-level reasoning by unifying unstructured (\eg, the context in question and answer; ``QA context'') and structured 
(\eg, knowledge graph for the QA context and scene; ``concept graph") multimodal knowledge.  

Most of the high-performing VQA methods~\cite{lu2019vilbert,Su2020VL-BERT,li2020oscar,yu2020ernie,chen2020uniter,zellersluhessel2021merlot,zellers2022merlot} pretrain a multimodal transformer model on a large-scale dataset to obtain unstructured multimodal knowledge from image and language contexts, and then finetune the pretrained model to reason on downstream tasks (\eg, visual commonsense reasoning (VCR) task \cite{zellers2019vcr}).
Existing methods (\eg, SGEITL~\cite{wang2021sgeitl}) also incorporate structured knowledge into these transformer-based models by including a scene graph in the input of a pretrained multimodal transformer model.
More recent methods~\cite{Marino_2021_CVPR,Zhang_2021_CVPR} further combine the scene graph and the concept graph by inter-connecting corresponding visual nodes and concept nodes through graph neural networks (GNNs), and then incorporate the unstructured QA context representation to perform question answering.  
However, these methods only perform late fusion or unidirectional fusion from unstructured knowledge to structured knowledge and do not train the model to mutually aggregate information from both sides. This can limit their potential to perform joint reasoning over the heterogeneous modalities of knowledge.
As unstructured knowledge and structured knowledge have complementary benefits---pretrained unstructured representations capture broader knowledge and structured representations offer scaffolds for reasoning---\cite{yasunaga2021qagnn}, this motivates the development of models that deeply fuse the two modalities of knowledge for visual question answering.

We propose VQA-GNN (Figure~\ref{fig:vqa_gnn_concept}), a new visual question answering model performing bidirectional fusion between unstructured and structured multimodal knowledge to obtain a unified, more expressive knowledge representation.
VQA-GNN extracts a scene graph from the given input image using an off-the-shelf scene graph generator~\cite{tang2020unbiased} and then retrieves a relevant concept graph for the input image and QA context from a general knowledge graph like ConceptNet ~\cite{Speer_Chin_Havasi_2017}, obtaining a structured representation of the scene. 
Simultaneously, to obtain an unstructured knowledge representation for the scene, (1) we use pretrained RoBERTa~\cite{liu2019roberta} to encode the context in question and answer (``QA-context") as \textit{QA-context node}, and (2) we retrieve relevant visual regions from a general scene graph VisualGenome~\cite{krishna2017visual} and take their mean pooled representation as a \textit{QA-concept node}, which we connect to the scene graph. 
We then connect the scene graph and the concept graph through \textit{QA-context node} to build a multimodal semantic graph.

To achieve bidirectional fusion across the multimodal semantic graph, we introduce a new multimodal GNN technique that performs inter-modal message passing. 
The multimodal GNN consists of two modality-specialized GNN modules, one for each modality, which perform inter-message aggregation between the \textit{QA-context node} and nodes in structured graphs, aiming to reduce representational gaps between modalities.
Meanwhile, by leveraging the robust transformer-based architecture of RoBERTa, we unfreeze and finetune the weights of the QA-context node to enable mutual information aggregation from modality-specialized GNN modules.

We evaluate VQA-GNN on two challenging VQA tasks, VCR~\cite{zellers2019vcr} and GQA~\cite{Hudson_2019_CVPR}. These tasks require systems to perform conceptual and compositional reasoning to answer diverse questions (\eg, multiple-choice question answering and rationale selection in VCR; open-domain question answering in GQA). 
Our model outperforms strong baseline VQA methods \cite{wang2021sgeitl,hu2019language} by \textbf{3.2\%} on VCR (Q-AR) and \textbf{4.6\%} on GQA. Moreover, ablation studies show the efficacy of our two main techniques, bidirectional fusion and multimodal GNN message passing. On VCR, our multimodal GNN technique that reduces multimodal gaps outperforms existing works that use generic GNNs \cite{Marino_2021_CVPR, Zhang_2021_CVPR} by \textbf{4.5\%}. On GQA, bidirectional fusion outperforms a unidirectional fusion variant by \textbf{4\%}. These results confirm the promise of \methodname in unifying unstructured and structured multimodal knowledge for reasoning.

\section{Problem Setup}\label{sec:problem}
This work focuses on multiple-choice and open-domain visual question answering, respectively. Each data point consists of an image $c$, and a natural language question $q$. For the multiple-choice setting, each question corresponds to a set of candidate answers $\mathcal{A}$, where only one candidate $a_\text{correct} \in \mathcal{A}$ is the correct answer to the question. Given a QA example $(c, q, \mathcal{A})$, we assume we have access to its relevant joint graph $\mathcal{G}^{(vcr)}$ and our goal is to identify the correct answer $a_\text{correct}\in\mathcal{A}$.
For the open-domain setting, all questions correspond to a large set of common answer classes $\mathcal{B}$, where only one candidate $b_\text{correct} \in \mathcal{B}$ is the best answer to each question. Given a data example $(c, q)$ with relevant scene graph $\mathcal{G}^{(gqa)}$, the goal is to identify $b_\text{correct} \in \mathcal{B}$. 
\section{Related Work}

\subsection{Multimodal transformer}
VQA has emerged as one of the most popular topics in
the computer vision community over the past few years \cite{VQA,fukui2016multimodal,ben2017mutan,kim2018bilinear,Lou_2022_CVPR,Hong_2021_CVPR}.
Existing methods for VQA \cite{lu2019vilbert,Su2020VL-BERT,li2020oscar,zellersluhessel2021merlot} employ the pretrain-and-finetune approach, where they train a multimodal transformer model on large-scale visual-language datasets, and then finetune the pretrained model on the downstream VQA datasets, \eg, RESERVE-L model \cite{zellers2022merlot} is pretrained using 1 billion image-caption data including video frames, text, and audio.
However, these methods only focus on obtaining unstructured multimodal representations by modeling implicit interactions over the visual and language domains.
In contrast, our method introduces a multimodal GNN module to obtain unified knowledge representations from unstructured and structured multimodal knowledge based on explicit interactions over a well-structured multimodal semantic graph. 

\subsection{Structured knowledge-based VQA}
\noindent\textbf{Scene graph.}
Existing methods such as~\cite{yu2020ernie} introduce a scene graph prediction task to learn structured knowledge conditioned multimodal representations, and the work~\cite{wang2021sgeitl} proposes to incorporate extracted scene graph in multimodal transformer models. 
These works \cite{teney2017graph,learningconditionedgraph,hu2019language,liang2021graghvqa} also exploit GNNs \cite{kipf2016semi,velickovic2018graph,xu2018how,busbridge2019rgat} to incorporate unstructured QA-context knowledge into a structured scene graph for question answering. However, these methods only perform late fusion or unidirectional fusion from unstructured knowledge to structured knowledge. In contrast, our method performs bidirectional fusion to unify unstructured and structured knowledge.

\noindent\textbf{Concept graph.}
Aiming to achieve concept-level reasoning beyond image-level recognition for visual understanding, existing works \cite{shah2019kvqa,singh2019strings,okvqa,Ziaeefard2020TowardsKV,garderes-etal-2020-conceptbert,krvqa,Marino_2021_CVPR,Zhang_2021_CVPR,Wu2022MultiModalAV,Li2022DynamicKM,Ding_2022_CVPR,yasunaga2022retrieval} utilize knowledge graphs (KGs) to explore how to unify commonsense knowledge \cite{yasunaga2021qagnn,yasunaga2022dragon,ren2021lego} about background concepts of the scene. The work \cite{Yang2021AnES} converts the image into captions and performs GPT-3 \cite{gpt3} in joint knowledge retrieval and reasoning. 
The work~\cite{Li2022DynamicKM} encodes question-related knowledge from the retrieved knowledge facts to a knowledge-aware question representation, and then performs a question and knowledge-guided graph attention operation for answer reasoning. However, structured concept knowledge relevant to the QA context is not enough to represent the background scene. We build a concept graph to cover structured and unstructured concept knowledge relevant to the QA context as well as the background scene.

\noindent\textbf{Scene graph \& concept graph.}
To enrich structured knowledge, these works \cite{Ziaeefard2020TowardsKV,garderes-etal-2020-conceptbert,Wu2022MultiModalAV} utilize GNNs to learn graph representations of the scene graph and concept graph respectively, and then perform later fusion across the QA context, scene graph and concept graph for question reasoning. However, it is insufficient to capture the interactions across different modalities for concept-level reasoning. These works \cite{Marino_2021_CVPR, Zhang_2021_CVPR} unify the scene graph and concept graph by interconnecting corresponding visual and concept nodes to capture their interactions. However, the representational gap between modalities adversely affects the performance of inter-modal message passing for capturing joint reasoning \cite{yanan_access_2022,liang2022mind}. Our method inter-connects the scene graph and concept graph via a QA context node and introduces a new multimodal GNN technique to mitigate representational gaps between modalities.

\section{Methodology}

\begin{figure*}[t]
\centering
\includegraphics[width=\linewidth]{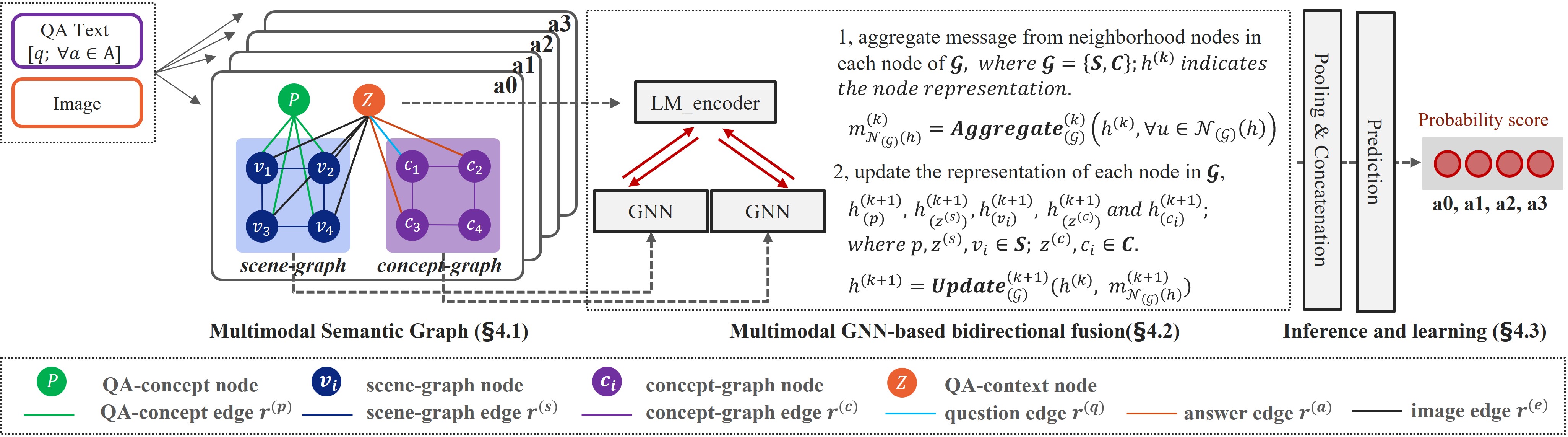}
\caption{Reasoning procedure of VQA-GNN. We first build a multimodal semantic graph for each given image-QA pair to unify unstructured (\eg, ``node p" and ``node z") and structured (\eg, ``scene-graph" and ``concept-graph") multimodal knowledge (\S \ref{semantic-graph}). Then we perform inter-modal message passing with a multimodal GNN-based bidirectional fusion method (\S \ref{gnn}) to update the representations of node $z$, $p$, $v_i$ and $c_i$ for $k+1$ iterations in two steps. Finally, we predict the answer with these updated various node representations (\S \ref{infer}).
Here, ``S" and ``C" indicate scene-graph and concept-graph respectively. ``LM\_encoder" indicates a language model used to finetune QA-context node representation, and ``GNN" indicates a relation-graph neural network for iterative message passing.}
\label{fig:vqagnn_architecture}
\end{figure*}

As shown in Figure \ref{fig:vqagnn_architecture}, given an image and its related question with an answer choice, first we build a multimodal semantic graph to unify unstructured and structured multimodal knowledge into a joint graph (\S \ref{semantic-graph}). Then we propose a multimodal GNN-based bidirectional fusion method that performs inter-modal message passing to obtain node representations enhanced with unstructured and structured multimodal knowledge (\S \ref{gnn}). Finally, we get the pooled representations of scene-graph and concept-graph and concatenate them with the representations from the QA-context node and QA-concept node for answer prediction (\S \ref{infer}).  

\subsection{Multimodal semantic graph} \label{semantic-graph}

\noindent\textbf{Scene-graph encoding.} 
Given an image, we use a pretrained scene graph generator to extract a scene graph that consists of recall@20 of $(subject, predicate, object)$ triplets to represent structured image context \cite{tang2020unbiased}, \eg, $(car, behind, man)$. 
Then we apply a pretrained object detection model for embedding a set of scene graph nodes $\mathcal{V}^{(s)}= \{v_i\}_{i=1}^{N}$ (N indicates the maximum number of scene-graph nodes of ``20") and represent $v_i^{(s)}$ with a $2048$ dimensional visual feature vector \cite{zhang2021vinvl}.
We indicate the predicate edge types in the scene graph with a set of scene graph edges $\mathcal{E}^{(s)}= \{r_i^{(s)}\}_{i=1}^{D}$ (D denotes the number of edge types) and represent $r_j^{(s)}$ with a $D$-dimensional one-hot vector.

\noindent\textbf{QA-concept node retrieval.}
In addition to the local image context, with an assumption that the global image context of the correct choice aligns with the local image context, we employ a pretrained sentence-BERT model to calculate the similarity between each answer choice and all descriptions of the region image within the VisualGenome dataset~\cite{krishna2017visual}. This process allows us to extract relevant region images that capture the global image context associated with each choice~\cite{reimers-2019-sentence-bert}.
We retrieve the top 10 results and utilize the same object detector to embed them. These embeddings are averaged to obtain a QA-concept node denoted as $p$. Subsequently, we introduce a QA-concept edge, denoted as $r^{(p)}$, which serves to fully connect node $p$ with node $v_i$.

\begin{figure}[h]
\centering
\scalebox{0.9}{
\includegraphics[width=\columnwidth]{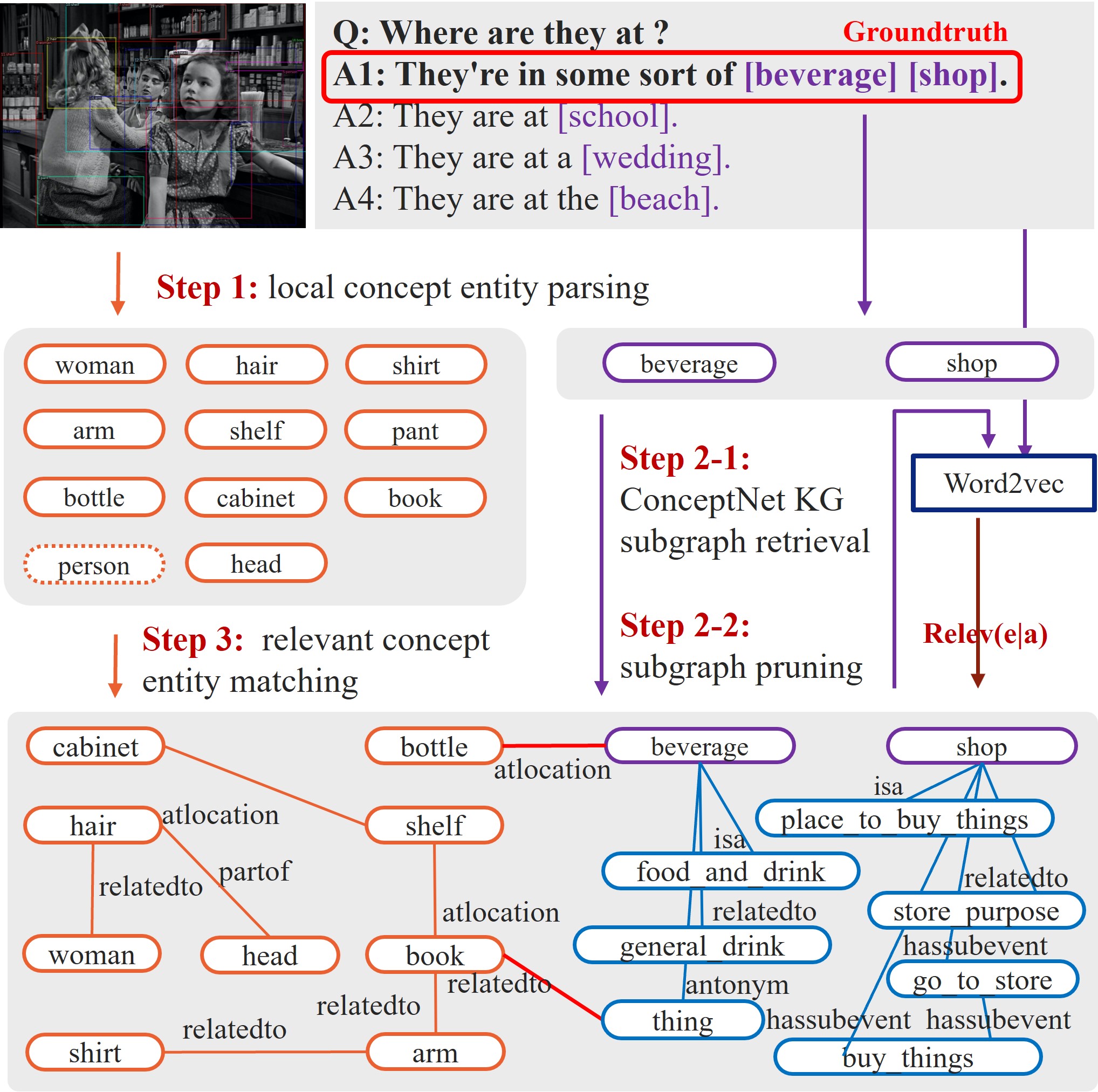}
}
\caption{The process of concept-graph retrieval involves the calculation of similarity between concept-graph nodes and the answer context, denoted as $Relev(e|a)$.}
\label{fig:concept-level-knowledge}
\end{figure}

\noindent\textbf{Concept-graph retrieval.} We retrieve a concept graph from the image and ConceptNet KG, a general-domain knowledge graph~\cite{Speer_Chin_Havasi_2017}.
Our process is illustrated in Figure \ref{fig:concept-level-knowledge}. In \textbf{Step 1}, we extract concept entities from both the image and the answer choices. Specifically, for the image, we consider the detected object names as potential contextual entities, while excluding general terms like ``person" to streamline the reasoning process.
For the answer choice, we ground phases if they are mentioned concepts in the ConceptNet KG, \eg, ``beverage" and ``shop". In \textbf{Step 2-1}, we use grounded phases to retrieve their 1-hop neighbor nodes from the ConceptNet KG. In \textbf{Step 2-2}, since many concept nodes retrieved are semantically irrelevant to the answer choice, we use a word2vec model released by the spaCy library\footnote{\url{https://spacy.io/}} to get relevance score between concept node candidates and answer choices, and prune irrelevance nodes when the relevance score is less than 0.6.
As a result, given an answer choice, we can retrieve a relevance subgraph from ConceptNet KG based on the relevance score. In $\textbf{Step 3}$, to better comprehend concept knowledge from the image as well, in addition to linking adjacent object entities in the ConceptNet KG domain, we also combine parsed local concept entities of the image with the retrieved subgraph.
For instance, considering that ConceptNet encompasses various types of local concept entities, if a local concept entity (\eg, ``bottle'') is found adjacent to a retrieved entity (\eg, ``beverage''), we build a new knowledge triple, \eg, (bottle, aclocation, beverage).
Finally, we can construct a concept graph to depict the  structured knowledge at the concept level. We obtain a collection of concept-graph nodes denoted as $\mathcal{V}^{(c)}=\{c_i\}_{i=1}^{N}$, where $N$ represents the maximum number of concept-graph nodes of $60$.
The concept entity $c_i$ is represented using a $1024$-dimensional text feature vector as the concept entity embedding in \cite{feng-etal-2020-scalable}. 
Additionally, we initialize a set of concept-graph edges denoted as $\mathcal{E}^{(c)}= \{r_i^{(c)}\}_{i=1}^{D}$, using $D$-dimensional one-hot vectors, where $D$ is the number of edge types in concept-graph.

\noindent\textbf{QA-context node encoding.}
To construct a multimodal semantic graph, we introduce an unstructured QA-context node denoted as $z$ to inter-connect the scene-graph and concept-graph using three additional relation types: the question edge $r^{(q)}$, the answer edge $r^{(a)}$, and the image edge $r^{(e)}$.
The image edge $r^{(e)}$ fully links node $z$ with $\mathcal{V}^{(s)}$, capturing the relationship between the QA context and relevant entities within the scene-graph.
The question edge $r^{(q)}$ and answer edge $r^{(a)}$ link node $z$ with the entities extracted from the question and the answer text, respectively, capturing the relationship between the QA context and the relevant entities within the concept-graph.
As a result, we construct a multimodal semantic graph $\mathcal{G}=\{S, C\}$ to provide a joint reasoning space, which includes two sub-graphs of scene-graph $S$ and concept-graph $C$, two super nodes of QA-concept node and QA-context node. Here, the QA-concept node is included in $S$ and the QA-context is included in $S$ and $C$ for performing inter-modal message passing in \S \ref{gnn}.
Especially, the QA-context node $z$ is assigned to not only learn unstructured discriminative representations by giving a Q and A text pairs but also to incorporate structured multimodal knowledge from scene-graph and concept-graph for effective VQA. As the transformer-based method is powerful for multimodal representation learning \cite{Su2020VL-BERT,lu2019vilbert}, we employ the RoBERTa LM \cite{liu2020roberta} as the encoder of QA-context node $z$ and finetune it with GNN modules to achieve bidirectional multimodal knowledge fusion (see Figure~\ref{fig:vqagnn_architecture}). 

\subsection{Multimodal GNN-based bidirectional fusion} \label{gnn}
To improve inter-modal message passing by avoiding directly aggregating neighborhood nodes that may be initialized in different modality domains, we propose a multimodal GNN-based bidirectional fusion method built by two relation-graph neural networks for scene-graph and concept-graph respectively (see \S \ref{para:modality_gap}). The relation-graph neural network is built on the Graph Attention Networks (GAT) \cite{velickovic2018graph} by introducing multi-relation aware message for attention-based message aggregation process to better capture multiple relation information.

The detail of the relation-graph neural network is as follows:
we have five node types: $\mathcal{T} = \{\bm Z, \bm P,\bm S, \bm C\}$ in the multimodal semantic graph and they indicate QA-context node $z$, QA-concept node $p$, scene-graph node $s$, question node $q$ and concept-graph node $c$. As relation edge representation $\bm r_{i,j}$ should capture relationship from node $i$ to node $j$ and difference of node types represents a special relation between neighborhood nodes, we first obtain node type embedding $u_i$, $u_j$ and then concatenate them with edge embedding $e_{ij}$ to generate multi-relation embedding $\bm r_{ij}$ from $i$ to $j$ by 
\begin{eqnarray}\label{eq:2}
   &\bm r_{ij} = f_r([e_{ij}||u_i||u_j])
\end{eqnarray}
where $u_i,u_j \in \{0,1\}^{|\mathcal{T}|}$ are one-hot vectors indicating the node types of $i$ and $j$, $e_{ij} \in \{0,1\}^{|\mathcal{R}|}$ is a one-hot vector indicating relation type of edge $(i,j)$. $||$ is the concatenation operation, and $f_r:\mathbb{R}^{|\mathcal{R}|+2|\mathcal{T}|}\rightarrow \mathbb{R}^{\mathcal{D}}$ is a 2-layer MLP. Based on multi-relation embedding $\bm r_{ij}$, the multi-relation aware message $\bm m_{ij}$ from $i$ to $j$ is computed by 
\begin{eqnarray}\label{eq:3}
   &\bm m_{ij} = f_m([\bm h^{(k+1)}_{i}||\bm r_{ij}])
\end{eqnarray}
where $f_m:\mathbb{R}^{2\mathcal{D}} \rightarrow \mathbb{R}^{\mathcal{D}}$ is a linear transformation. $h^{(k+1)}_i$ is the node representation of each node $i$ in the graph. We then 
recursively updated it $k+1$ times by
\begin{eqnarray}\label{eq:4}
   & \bm h^{(k+1)}_i = f_h \left( \displaystyle \sum_{j \in \mathcal{N}_i}\alpha_{ij} \bm m_{ij} \right) + \bm h^{(k)}_i
\end{eqnarray}
where $f_h:\mathbb{R}^{\mathcal{D}} \rightarrow \mathbb{R}^{\mathcal{D}}$ is 2-layer MLP with batch normalization \cite{bn}. $\mathcal{N}_i$ indicates the neighborhood of node $i$, $\alpha_{ij}$ is an attention weight to emphasize important messages passed from $\mathcal{N}_i$ to node $i$. We obtain $\bm q_i,\bm k_j$ by 
\begin{eqnarray}\label{eq:5}
   & \bm q_i = f_q(\bm h^{(k+1)}_{i}), \bm k_j = f_k([\bm h^{(k+1)}_{j}||\bm r_{ij}])
\end{eqnarray}
where $f_q:\mathbb{R}^{\mathcal{D}} \rightarrow \mathbb{R}^{\mathcal{D}}$ and $f_k:\mathbb{R}^{2\mathcal{D}} \rightarrow \mathbb{R}^{\mathcal{D}}$ are linear transformations. $\alpha_{ij}$ is computed using the softmax function by
\begin{eqnarray}\label{eq:6}
   & \gamma_{ij} = \frac{\bm q^T_i \bm k_j}{\sqrt{D}}, \\
   & \alpha_{ij} = \operatorname{softmax}_{j}(\gamma_{ij}) = \displaystyle \frac{\exp({\gamma_{ij}})}{\sum_{j' \in \mathcal{N}_i}\exp({\gamma_{ij'}})}
\end{eqnarray}
By referring to Eq. \ref{eq:4}, we perform message passing to update node representations in each graph in parallel by aggregating multi-relation aware messages from neighborhood nodes in each node. As a result, we obtain structured graph node representations $h^{(k+1)}_{(v_i)}$ and $h^{(k+1)}_{(c_i)}$, unstructured node representations $h^{(k+1)}_{(p)}$ and $h^{(k+1)}_{(z)}$. For node $z$, we update it with scene-graph and concept-graph respectively, and concatenated by  
\begin{eqnarray}\label{eq:7}
  & \bm h^{(k+1)}_{(z)} = f_z([\bm h^{(k+1)}_{(z^{(s)})}||\bm h^{(k+1)}_{(z^{(c)})}])
\end{eqnarray}
where $f_z:\mathbb{R}^{2\mathcal{D}} \rightarrow \mathbb{R}^{\mathcal{D}}$ is a linear transformation.

\subsection{Inference and Learning} \label{infer}
To identify the correct answer $a_{correct} \in \mathcal{A}$ with a QA example $(c,q,\mathcal{A})$, we compute the probability $p(a|c,q)$ for each answer choice with its multimodal semantic knowledge from scene-graph, concept-graph, QA-context node, and QA-concept node. With various node representations on the \textit{L-th} ($L=k+1$) layer updated by GNN modules (shown in Figure \ref{fig:vqagnn_architecture}), we obtain pooling representations $\bm h^{(K+1)}_{(s)}$ and $\bm h^{(K+1)}_{(c)}$ of scene-graph and concept-graph and then concatenate with QA-context node and QA-concept node representations. Finally we calculate $p(a|c,q)$ by
\begin{eqnarray}\label{eq:8}
   & \displaystyle \bm h^{(k+1)}_a = [\bm h^{(K+1)}_{(s)} || \bm h^{(K+1)}_{(c)} || \bm h^{(K+1)}_{(p)}|| \bm h^{(K+1)}_{(z)}], \\
   & \operatorname{logit}(a) = f_c(\bm h^{(k+1)}_a),\\
   & p(a|c,q) = \operatorname{softmax}_a(\operatorname{logit}(a)) 
\end{eqnarray}
where $\operatorname{logit}(a)$ indicates the confident score of answer choice $a$, $f_c:\mathbb{R}^{4\mathcal{D}} \rightarrow \mathbb{R}^{1}$ is a linear transformation that maps the concatenation of representations to a scale. We normalize it across all answer choices using the softmax function. 
For the training process, we apply the cross entropy loss to optimize the \textit{\methodname} model end-to-end.

\section{Experiments}\label{sec:experiment}

\begin{table*}[!t]
\centering
\resizebox{\linewidth}{!}{%
\begin{tabular}{lcccccccc}  
\toprule
\multirow{2}{*}{Model} & \# Image-caption & \multirow{2}{*}{Parameters} & \multirow{2}{*}{Structured knowledge} & \multicolumn{3}{c}{Test Acc.(\%)} \\ 
\cmidrule(lr){5-7}

&in pretraining& &  & Q$\rightarrow$A& QA$\rightarrow$R & Q$\rightarrow$AR \\
\midrule
ViLBERT \cite{lu2019vilbert} &3.3M&\textbf{221M}&No&73.3 & 74.6 & 54.8 \\
VLBERT-L \cite{Su2020VL-BERT} &3.3M & 383M &No&75.8 & 78.4 & 59.7 \\
SGEITL+VLBERT \cite{wang2021sgeitl}& 290k & $\geq$ 383M & Yes & 76.0 & 78.0 & 59.6\\
UNITER-(B/L)\cite{chen2020uniter} & 9.5M& 154M/378M &No&75.0/77.3&77.2/80.8&58.2/62.8\\
ERNIE-ViL-(B/L) \cite{yu2020ernie} &3.8M& 212M/533M &No&77.0/79.2&80.3/83.5&62.1/66.3\\
\midrule
\textbf{VQA-GNN (Ours)} &\textbf{290k} &372M  & Yes  &77.9 & 80.0 & 62.8 \\
\midrule
MERLOT \cite{zellersluhessel2021merlot}& 180M&223M &No&80.6&80.4&65.1 \\
RESERVE-(B/L) \cite{zellers2022merlot} & 1B& 200M/644M&No &79.3/84.0& 78.7/84.9 &62.6/72.0\\
RESERVE-L + \textbf{VQA-GNN (Ours)} &1B &1B  & Yes  & \textbf{85.3} & \textbf{86.9} & \textbf{74.3} \\

\bottomrule
\end{tabular}
}
\caption{Accuracy scores for VCR test set. 
\textit{\methodname} outperforms \textit{SGEITL+VLBERT} model on Q$\rightarrow$AR metric by \textbf{3.2\%}, and achieves competitive accuracy with SOTA methods, which have a close number of parameters but SOTA methods require a large amount of image caption data in pre-training process (over 13x larger than our model), \eg, \textit{"UNITER-L"}, \textit{"ERNIE-ViL-B"}, \textit{"RESERVE-B"}. Moreover, \textit{"RESERVE-L+VQA-GNN"} outperforms \textit{RESERVE-L} by \textbf{2.3\%} on Q$\rightarrow$AR metric.} 
\label{tab:table1}
\end{table*}

\subsection{Experiment Setup} \label{sec:trainsetting}
\noindent\textbf{Visual Commonsense Reasoning (VCR).} \label{sec:trainsetting_vcr}
We evaluate VQA-GNN on VCR \cite{zellers2019vcr}. It contains 290k pairs of questions, answers, and rationales, over 110k unique movie scenes. VCR consists of two tasks: visual question answering (Q$\to$A), answer justification (QA$\to$R). Each question in the dataset is provided with four candidate answers. The goal of (Q$\to$A) is to select the best answer, while the goal of (QA$\to$R) is to justify the given question answer pair by picking the best rationale out of the four candidates. We joint train \textit{\methodname} on Q$\rightarrow$A and QA$\rightarrow$R, with a common LM encoder, the multimodal semantic graph for Q$\rightarrow$A, a concept graph retrieved by giving question-answer pair with a rationale candidate for QA$\rightarrow$R.
We use a pretrained RoBERTa Large model to embed the QA-context node, and finetune it with the multimodal GNN for $50$ epoch by using learning rates $1$e-$5$ and $1$e-$4$ respectively. We set the number of layers ($L=5$) of VQA-GNN and use AdamW \cite{Kingma2015AdamAM} optimizer to minimize the loss. We use a linear warmup of the learning rate over the \textit{15-th} epoch, with a cosine decay thereafter to \textit{0}.

\noindent\textbf{GQA dataset.}  \label{sec:gqa_evaluating}
It contains open-ended questions (1.5M questions correspond to 1,842 answer tokens), along with 110K scene graphs and the semantic functional programs to offer unambiguous instructions \cite{Hudson_2019_CVPR}. We only use questions without giving a semantic feature program that limits the development of the model's reasoning abilities in a more practical setting. We define the question as the context node (node q) to fully connect visual and textual scene graphs (SG) respectively to structure multimodal semantic graphs. The node q is embedded with a pretrained RoBERTa large model, and we initialize object nodes' representations in visual SG using official object features, object nodes in textual SG by concatenating GloVe \cite{pennington-etal-2014-glove} based word embedding of the object name and attributes. Different from the training target of VCR, the goal of GQA is to classify the given image-question pair out of 1,842 answer classes. We finetune the node q with VQA-GNN for $50$ epoch by using learning rates $2$e-$5$ and $2$e-$4$ respectively.

\subsection{Performance}
\subsubsection{Evaluation on VCR dataset}
\noindent\textbf{Comparison with state-of-the-art methods.}
We compared \textit{\methodname} with state-of-the-art methods on the VCR test set in Table \ref{tab:table1}. Compared with the unidirectional fusion method \textit{SGEITL+VLBERT} that can boost multimodal transformer model VLBERT by incorporating visual scene graphs, \textit{\methodname} is a multimodal GNN-based bidirectional fusion method built on the multimodal semantic graph. Both were not pretrained on the large-scale dataset. 
\textit{\methodname} improves SGEITL+VLBERT on the Q$\rightarrow$AR metric by \textbf{3.2\%}, and further reduces over \textbf{11M} training parameters. 
We think that the structured multimodal semantic graph provides much more commonsense knowledge related to QA and original image than SGEITL, and the multimodal GNN-based bidirectional fusion method works much better on unifying unstructured and structured multimodal knowledge than multimodal transformer models. Moreover, since we retrieve commonsense knowledge from structured multimodal semantic graphs directly, \textit{\methodname} is a cost-effective approach compared to multimodal transformer models that consume much GPU resources to learn commonsense knowledge with large parameters.

We also demonstrate the effectiveness of \textit{\methodname} by comparing it with state-of-the-art multimodal transformer models that were pretrained across text and images and were finetuned on the VCR dataset. As shown in Table \ref{tab:table1}, the larger image caption data and parameters, the higher performance the model can achieve. In contrast, \textit{\methodname} trained with VCR dataset with $290K$ image-caption pairs performs similarly to UNITER-L that requires over \textbf{32x} larger image-caption data than us in pretraining process. These results suggest that \textit{\methodname} obtaining structured context knowledge inferred from image-level and concept-level knowledge sources is as effective as the pretraining process for previous methods. 
Moreover, \textit{\methodname} can further enhance RESERVE-L performance on both Q$\rightarrow$A and QA$\rightarrow$R, and finally improves the score by \textbf{2.3\%} on Q$\rightarrow$AR metric. As correcting some questions requires the model to understand commonsense knowledge related to image context and have good reasoning ability, it is difficult for multimodal transformer methods including RESERVE-L.  
On the other hand, \textit{\methodname} not only structures a joint semantic graph to provide commonsense knowledge related to image context but also has a good reasoning ability thanks to its multimodal GNN module.
Additionally, in the supplementary material, we detail the results compared to baselines pretrained only with the VCR dataset, as well as the evaluation of different question types.

\noindent\textbf{Effectiveness of the multimodal semantic graph.}
To further study the behavior of modules in the multimodal semantic graph, and quantitatively evaluate pretrained models used in this work (\eg, RoBERTa-L, scene-graph[scene graph generator], concept-graph[conceptNet KG]), we report the performance of using different node representations in Table \ref{tab:semantic_graph}. 
We respectively build classification models by applying Node p and Node z to get their validation accuracy on Q$\rightarrow$A subtask. The scene-graph structured by connecting Node p and Node z with extracted visual scene graph improves over $25\%$ on average of these two nodes. In terms of concept-graph, it is structured by connecting Node z with retrieved conceptual triplets from ConcepNet KG, improving Node z's performance by $15.2\%$. We further compare \textit{\methodname} on ``scene-graph + concept-graph" w/ and w/o Node $p$, and the result shows that including Node p can further improve the performance by $2\%$. We believe that the Node p representing global visual knowledge associated with the correct answer is able to pass visual commonsense knowledge to the multimodal semantic graph, and it is effective besides employing ConcepNet KG to obtain textual commonsense knowledge \cite{yasunaga2021qagnn}.
\vspace{5mm}

\begin{table}[ht]
\centering
\scalebox{0.8}{
\begin{tabular}{lcc}  
\toprule
Model & Val Acc.(\%) (Q$\rightarrow$A)  \\
\midrule
Node p (Vinvl) &  43.5 \\
Node z (RoBERTa-L) &  53.8 \\
concept-graph & 69.0  \\
scene-graph & 73.7 \\
\midrule
concept-graph + scene-graph (w/o node $p$) & 75.1 \\
concept-graph + scene-graph (w/ node $p$) & \textbf{77.1} \\
\bottomrule
\end{tabular}
}
\caption{All modules in the multimodal semantic graph help boost the final performance. Here, ``scene-graph" includes node z and node p, ``concept-graph" includes node z.}
\label{tab:semantic_graph}
\end{table}
\vspace{-0.5cm}
\begin{table}[h]
\centering
\scalebox{0.8}{
\begin{tabular}{lc}  
\toprule
Model & Val Acc.(\%) (Q$\rightarrow$A)  \\
\midrule
Ablation 1 (single GNN) & 73.0 \\
Ablation 2 (single GNN w/ cross-modal edges) & 70.6 \\
\midrule
VQA-GNN (two modality-specialized GNNs) & \textbf{75.1} \\
\bottomrule
\end{tabular}
}
\caption{Ablation 1 and Ablation 2 indicate a single GNN on the multimodal semantic graph w/o and w/ direct cross-modal edges, respectively (Figure \ref{fig:ablation_1}). VQA-GNN with two modality-specialized GNNs on the multimodal semantic graph achieves the best score.}
\label{tab:multi_gnn}
\end{table}

\begin{figure}[h]
\centering
\includegraphics[width=\columnwidth]{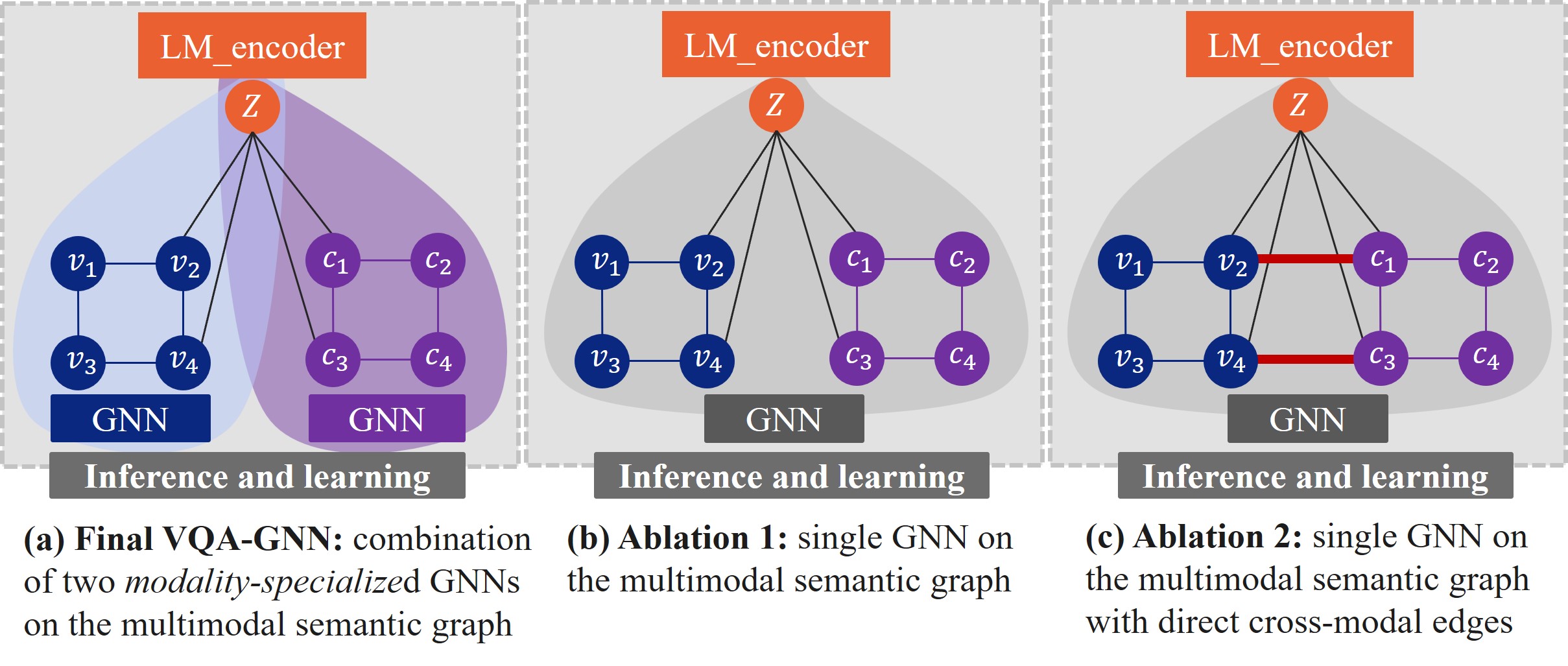}
\caption{Ablation architectures. We find that our final VQA-GNN architecture with two modality-specialized GNNs overcomes the representation gaps between modalities (\S \ref{para:modality_gap}).}
\label{fig:ablation_1}
\end{figure}

\noindent\textbf{Analysis of the multimodal GNN method.} \label{para:modality_gap}
To analyze the effect of the multimodal GNN method on mitigating the multimodal gap in performing inter-modal message passing, we compared the final VQA-GNN with two single GNNs built on multimodal semantic graphs with and without direct cross-modal edges in Figure \ref{fig:ablation_1}. As the results of VCR validation set shown in Table \ref{tab:multi_gnn}, the final VQA-GNN built with the multimodal GNN on the multimodal semantic graph improves the accuracy of both ablative architecture by over \textbf{2\%}. We believe that the multimodal GNN built by two modality-specific GNNs can effectively avoid directly aggregating nodes from scene-graph and concept-graph to alleviate the modality gap. As a result, the inter-modal message passing can be improved. We further explored the aggregation process for some node samples to demonstrate why the two ablation architectures fail to alleviate the modality gap. Here, $m^{(k)}_{\mathcal{N}(u)}$ represents the aggregated messages from all neighbors of node $u$ at the $k$-th iteration.
\begin{eqnarray}\label{eq:exp_1}
   & \displaystyle m^{(k)}_{\mathcal{N}(u)} = \operatorname{Aggregate^{(k)}}(u^{(k)}, \forall v \in \mathcal{N}(u))
   \end{eqnarray}
where $\mathcal{N}(u)$ denotes a set of neighborhood nodes of the node $u$, and $k$ denotes the iterations of $m^{(k)}_{\mathcal{N}(u)}$.

For (c) Ablation 2 in Figure \ref{fig:ablation_1}, we assume that node $v_2$ is connected with node $c_1$ as both represent the same notion. However, their feature vectors are distributed in different modality domains and affect the aggregation process. We show the neighborhood nodes of QA-context node $z$, visual node $v_2$ and concept node $c_1$ are follows:
\begin{eqnarray}\label{eq:exp_2}
   & \displaystyle \mathcal{N}(z) = \{v_2,v_4,c_1,c_3\}\\
   & \displaystyle \mathcal{N}(v_2) = \{z,v_1,v_4,c_1\}; \mathcal{N}(c_1) = \{z,c_2,c_3,v_2\}
\end{eqnarray}
where their neighborhood nodes include heterogeneous nodes from different modality domains.

For (b) Ablation 1 in Figure \ref{fig:ablation_1}, the neighborhood nodes of QA-context node $z$, visual node $v_2$ and concept node $c_1$ are follows:
\begin{eqnarray}\label{eq:exp_3}
   & \displaystyle \mathcal{N}(z) = \{v_2,v_4,c_1,c_3\}\\
   & \displaystyle \mathcal{N}(v_2) = \{z,v_1,v_4\}; \displaystyle \mathcal{N}(c_1) = \{z,c_2,c_3\}
\end{eqnarray}
Compared with (c) Ablation 2, node $c_1$ and node $v_2$ are removed from the neighborhood nodes of $v_2$ and $c_1$ which helped improve the performance of (c) Ablation 2 by 2.4\%. However, it is limited by the QA-context node $z$ that aggregates messages across scene-graph and concept-graph. Although QA-context node $z$ is a pretrained LM that can be finetuned on multimodal domains, it is more difficult to adapt to two modalities (Eq. \ref{eq:exp_3}) than to a single modality (Eq. \ref{eq:exp_4}). 
In contrast, the multimodal GNN method is designed by introducing two GNNs for each modality. We perform aggregation for QA-context node $z$ for each modality so that the pretrained LM is finetuned on a single modality to alleviate the modality gap. The neighborhood nodes of QA-context node $z$, visual node $v_2$ and concept node $c_1$ are follows:
\begin{eqnarray}\label{eq:exp_4}
   & \displaystyle \mathcal{N}(z)^{(m1)} = \{v_2,v_4\}; \mathcal{N}(z)^{(m2)} = \{c_1,c_3\}\\
   & \displaystyle \mathcal{N}(v_2) = \{z^{(m1)},v_1,v_4\}; \mathcal{N}(c_1) = \{z^{(m2)},c_2,c_3\}
\end{eqnarray}
where $m1$ and $m2$ indicate two message passing methods for each modality.

\subsubsection{Evaluation on GQA dataset}
\noindent\textbf{Comparison with baselines.} We also compared \textit{\methodname} with baseline models on GQA dataset, under the realistic setup of not using the annotated semantic functional programs (see \S \ref{sec:trainsetting}). As the results shown in Table \ref{tab:gqa_eva}, our model achieves validation accuracy of 58.9\% for visual SG and 87.9\% for textual SG. Compared with SGEITL \cite{wang2021sgeitl} and GCN \cite{liang2021graghvqa} which are unidirectional fusion methods, our method performs bidirectional fusion to unify unstructured and structured knowledge, and improved the reasoning ability of SGEITL by \textbf{5.6\%} and GCN by \textbf{2.2\%}. Moreover, by inter-connecting the visual and textual SG, our method achieves validation accuracy of \textbf{90.3\%} and further suggests its efficacy in performing inter-modal message passing.  

\begin{table}[h]
\centering
\scalebox{0.8}{
\begin{tabular}{lccccc}  
\toprule
Model &  Visual SG & Textual SG & Val Acc.(\%) \\
\midrule
SGEITL\cite{wang2021sgeitl} & \checkmark &  & 53.3 \\
CFR\cite{nguyen2022coarse}  &\checkmark & \checkmark & 73.6 \\
GCN\cite{liang2021graghvqa}&  & \checkmark & 85.7 \\
\midrule
 &\checkmark &  & 58.9 \\
VQA-GNN &  & \checkmark & 87.9 \\
 &\checkmark & \checkmark & \textbf{90.3} \\
\bottomrule
\end{tabular}
}
\caption{Accuracy scores on the GQA validation set. All models are trained under the realistic setup of not using the annotated semantic functional programs.}
\label{tab:gqa_eva}
\end{table}


\begin{table}[h]
\centering
\captionsetup[table]{skip=5pt}
\scalebox{0.8}{
    \begin{tabular}{lccc}
    \toprule
    Method & Val Acc.(\%) $\uparrow$ & Inference time (ms) $\downarrow$ \\
    \midrule
    Average pooling & 62.3 ($\pm 0.40$) & \textbf{5.2} &\\
    Unidirectional fusion & 86.3 ($\pm 0.01$) & 8.6\\
    Bidirectional fusion \textbf{(ours)} & \textbf{90.3} ($\pm 0.03$) & 5.5\\
    \bottomrule
    \end{tabular}
}
\caption{Ablation results on the effect of our proposed bidirectional fusion for GQA.}
\label{tab:gqa_ablation_result}
\end{table}

\begin{figure}[h]
\centering
\includegraphics[width=\columnwidth]{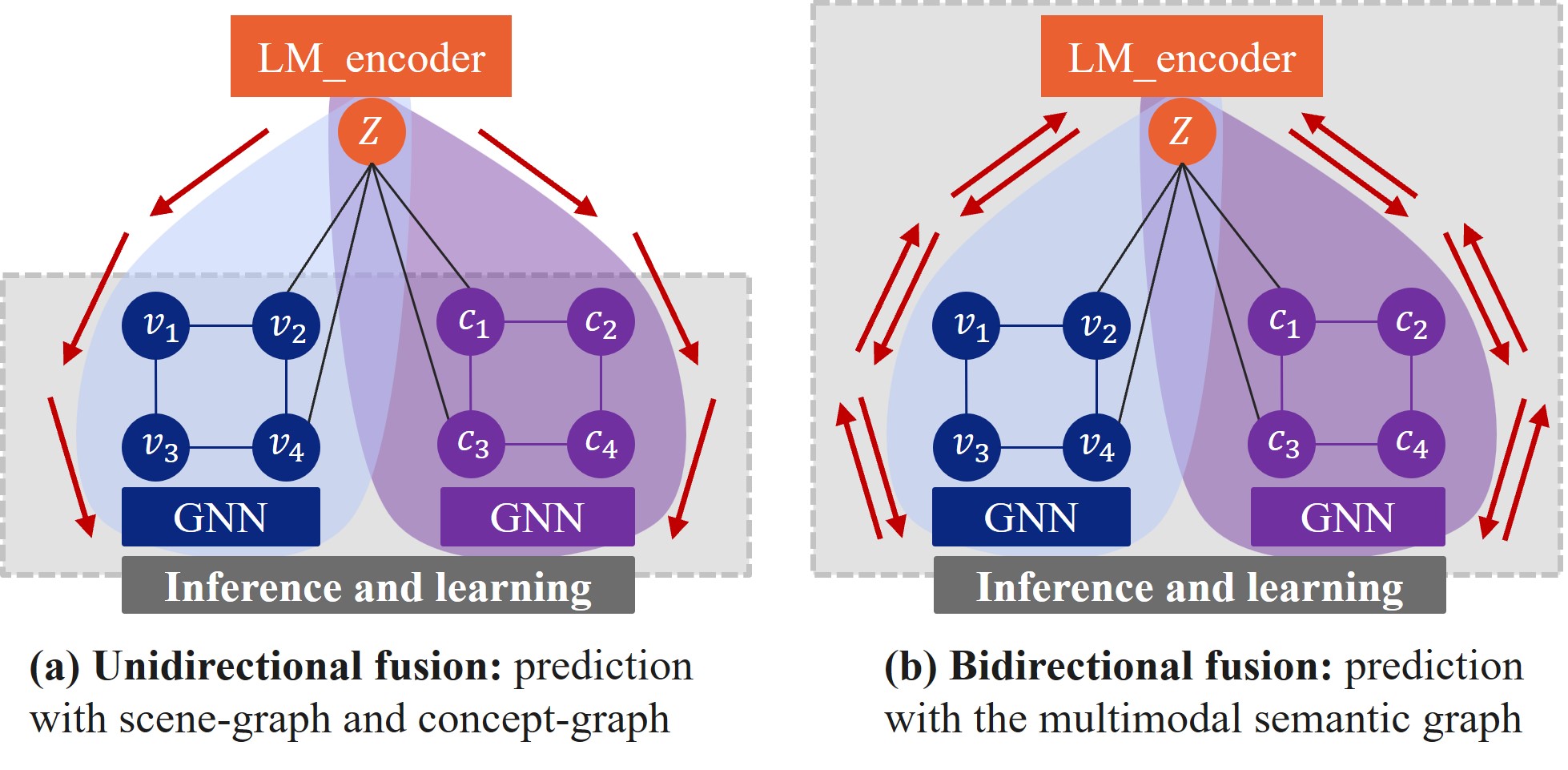}
\caption{Illustration of two knowledge fusion methods: our proposed bidirectional fusion v.s. the unidirectional fusion baseline.}
\label{fig:ablation_2}
\end{figure}

\noindent\textbf{Ablation study on the bidirectional fusion.}
To fairly study the effect of bidirectional fusion for improving concept-level reasoning, we evaluated the performance of \textit{\methodname} with and without structured multimodal knowledge-enhanced question representations. We show their difference in Figure \ref{fig:ablation_2}, compared with the unidirectional fusion, the bidirectional fusion approach is able to utilize the message aggregated from scene-graph and concept-graph in node $z$ to predict the correct answer. It facilitates the joint reasoning ability of VQA-GNN in capturing bidirectional interactions between unstructured node $z$ and structured multimodal semantic graph. As a result in Table \ref{tab:gqa_ablation_result}, the bidirectional fusion approach further improved the performance of the unidirectional fusion approach by \textbf{4\%}. We also compared our approach with an average pooling method that simply averages all node representations. We indeed find that this ablation performs significantly worse than others, which suggests that our approach can capture special relationship information between different nodes but average pooling cannot.

\section{Conclusion} \label{conclusions}
We proposed a novel visual question answering method, \textit{\methodname}, which unifies unstructured and structured multimodal knowledge to perform joint reasoning of the scene. 
In the evaluation of two challenging VQA tasks (VCR and GQA), our method substantially outperforms existing models without pretraining using massive image-caption data under the same training setting, our method outperforms strong baseline VQA methods by \textbf{3.2\%} on VCR (Q-AR) and \textbf{4.6\%} on GQA, suggesting its strength in performing concept-level reasoning. Ablation studies further demonstrate the efficacy of the bidirectional fusion and multimodal GNN method in unifying unstructured and structured multimodal knowledge. 
In the next step, we will extend our work to the video domain and focus on obtaining temporal semantic knowledge to enhance the machine's reasoning ability.

\section*{Acknowledgment}
We thank Rok Sosic, members of the Stanford SNAP group, as well as our anonymous reviewers for valuable feedback.
We gratefully acknowledge the support of DARPA under Nos. HR00112190039 (TAMI), N660011924033 (MCS); Funai Foundation Fellowship; Microsoft Research PhD Fellowship; Masason Foundation Fellowship; Apple PhD Fellowship;
ARO under Nos. W911NF-16-1-0342 (MURI), W911NF-16-1-0171 (DURIP); NSF under Nos. OAC-1835598 (CINES), OAC-1934578 (HDR), CCF-1918940 (Expeditions), IIS-2030477 (RAPID),
NIH under No. R56LM013365;
Stanford Data Science Initiative, 
Wu Tsai Neurosciences Institute,
Chan Zuckerberg Biohub,
Amazon, JPMorgan Chase, Docomo, Hitachi, Intel, KDDI, Toshiba, NEC, Juniper, and UnitedHealth Group.




{\small
\bibliographystyle{ieee_fullname}
\bibliography{egbib}
}

\end{document}